\documentclass{article}
\usepackage{spconf,amsmath,graphicx}
\usepackage{times}
\usepackage{epsfig}
\usepackage{amsfonts}
\usepackage{amssymb}
\usepackage{xcolor}
\usepackage{multirow}
\usepackage[breaklinks=true,bookmarks=false]{hyperref}

\newif\ifrems
\ifrems
\newcommand{\remSylv}[1]{\textcolor{red}{\sout{#1}}}
\newcommand{\addSylv}[1]{\textcolor{blue}{#1}}
\newcommand{\comSylv}[1]{\textcolor{red}{\textit{#1}}}
\newcommand{\addmoi}[1]{\textcolor{green}{#1}}

\newcommand{\remsido}[1]{\textcolor{magenta}{\textit{#1}}}
\else
\newcommand{\remSylv}[1]{\textcolor{red}{\sout{}}}
\newcommand{\addSylv}[1]{{#1}}
\newcommand{\comSylv}[1]{\textcolor{red}{\textit{}}}
\newcommand{\addmoi}[1]{{#1}}

\newcommand{\remsido}[1]{\textcolor{magenta}{\textit{}}}
\fi
\newcommand{\NFA}{\textsc{NFA}}

\title{\textit{A contrario} Paradigm for YOLO-based Infrared Small Target Detection }
\name{Alina Ciocarlan$^{1,2}$, Sylvie Le Hegarat-Mascle$^{2}$, Sidonie Lefebvre$^{1}$, Arnaud Woiselle$^{3}$, Clara Barbanson$^{3}$}

\address{\textsuperscript{1}DOTA and LMA2S, ONERA, Université Paris-Saclay, F-91123 Palaiseau, France\\
\textsuperscript{2}SATIE Université Paris-Saclay, 91405 Orsay, France\\
\textsuperscript{3}Safran Electronics $\&$ Defense F-91344 Massy, France\\}
\begin{document}
%
\maketitle
\begin{abstract}
Detecting small \addSylv{to tiny} targets in infrared images is a challenging task in computer vision, especially when it comes to differentiating these targets from noisy or textured backgrounds. Traditional object detection methods such as YOLO struggle to detect tiny objects compared to segmentation neural networks, resulting in weaker performance when detecting small targets. To reduce the number of false alarms while maintaining a high detection rate, 
we introduce an \textit{a contrario} decision criterion into the training of a YOLO detector. The latter takes advantage of the \textit{unexpectedness} of small targets to discriminate them from complex backgrounds. Adding this statistical criterion to a YOLOv7-tiny bridges the performance gap between state-of-the-art segmentation methods for infrared small target detection and object detection networks. It also significantly increases the robustness of YOLO towards few-shot settings.
\end{abstract}
\section{Introduction}
\label{intro}
Detecting small objects in infrared (IR) images accurately is essential in various applications, including medical or security fields. Infrared small target detection (IRSTD) is a great challenge in computer vision, where the difficulties are mainly raised by (i)~the size of the targets (area below $20$ pixels), (ii)~ the complex and highly textured backgrounds, leading to many false alarms, \addSylv{and (iii)~}the learning \addSylv{conditions, namely} learning from small, little diversified and highly class-imbalanced datasets\addSylv{, since the number of target class pixels is very small} in comparison with the background class \addSylv{one}. 
The rise of deep learning methods has led to impressive \addSylv{advances} in object detection in the past decades, mostly thanks to their ability to learn from a huge amount of annotated data to extract non-linear features well adapted to the final task. 
\addSylv{F}or IRSTD\addSylv{,} semantic segmentation neural networks (NN) \addSylv{are the most widely used~\cite{kou2023infrared}}. These include ACM~\cite{dai2021asymmetric}, LSPM~\cite{huang2021infrared} and one of the recent state-of-the-art (SOTA) method, namely DNANet~\cite{li2022dense}, which consists of several nested UNets and a multiscale fusion module that enable the segmentation of small objects with variable sizes. However, a major issue of relying on segmentation NN for object detection is that object fragmentation can occur when tuning the threshold used to binarize the segmentation map. This can lead to many undesired false alarms and distort counting metrics. Object detection algorithms like Faster-RCNN~\cite{ren2015faster} or YOLO~\cite{redmon2016you} reduce this risk by explicitly localizing \addSylv{the} objects thanks to the bounding box regression. 
However, they often have difficulty in detecting tiny objects. Very few studies have focused on adapting such detectors for IRSTD~\cite{mou2023yolo}, and no rigorous comparison was made with SOTA IRSTD methods.

In this paper, we propose a novel YOLO detection head, called OL-NFA (for Object-Level Number of False Alarms), that is specifically designed for small object detection. This module integrates an \textit{a contrario} decision criterion that guides the feature extraction so that \textit{unexpected} objects stand out against the background and are detected. It is used to re-estimate the objectness scores computed by a YOLO backbone, and has been carefully implemented to allow the back-propagation during training. One advantage of using \textit{a contrario} paradigm is that it focuses on modeling the background, for which we have a lot of samples, rather than the objects themselves. In this way, the problems of class imbalance and little training data are bypassed by carrying out the detection by rejecting the hypothesis of the background distribution.
Our main contributions are as follows:
\begin{enumerate}
    \item We design a novel YOLO detection head that integrates \textit{a contrario} criterion for estimating the objectness scores. By focusing on modeling the background rather than the object itself, we relax the constraint of having lots of training samples.
    \item We compare both SOTA segmentation NN and object detection methods on a famous IRSTD benchmark and show that adding OL-NFA to a YOLOv7-tiny backbone bridges the performance gap between object detectors and SOTA segmentation NN for IRSTD.  
    \item On top of that, our method improves YOLOv7-tiny performance by a large margin ($39.2\%$ AP for 15-shot) in few-shot settings, demonstrating the robustness of the \textit{a contrario} paradigm in weak training conditions. 
\end{enumerate}


\section{Related work}

\subsection{\textit{A contrario} reasoning}
\label{sec:nfa_formulation}

\textit{A contrario} decision methods allow to automatically derive a decision criterion with regards to \addSylv{a} hypothesis test\addSylv{. They} draw inspiration from theories of perception, in particular the Gestalt theory~\cite{desolneux2007gestalt}. They consist in rejecting a naive model characterizing a destructured background by \addSylv{using an interpretable} detection threshold. The latter \addSylv{allows us} to control the Number of False Alarms (NFA), often defined as the product between the total number of tested \textit{objects} and the tail distribution of the law followed by the chosen naive model. \addSylv{An NFA value can then be associated to any given \textit{object} since the computed tail value depends on the object features.}
Several \textit{a contrario} formulations have been proposed in the literature. They depend on whether we consider grey level or binary images. In the first case, the most commonly used naive model is the Gaussian distribution of the pixel grey-level values~\cite{Robin2010, ipol.2019.263, vidal2019aggregated}. 
The latter has been integrated into a deep learning framework by~\cite{ciocarlan2023deep}, and has shown great performance for small target segmentation. 
In the second case, the most widely used naive model is the uniform spatial distribution of ``true'' pixels in the image grid. This leads to a binomial distribution of parameter $p$ for the number of ``true'' pixels $\kappa$ falling within any given parametric shape of area $\nu$ ~\cite{desolneux2003grouping, HegaratMascle2019}: 
\begin{equation}
\label{eq:nfa}
\NFA\left(\kappa,\nu,p\right)=\eta \sum_{i=\kappa}^{\nu}\binom{\nu}{i} p^{i}\left(1-p\right)^{\nu-i},
\end{equation}
where $\eta$ is the number of tested objects. 
\addSylv{Based on Eq.~\eqref{eq:nfa}, a subset of pixels likely to represent an object is all the more significant as it contains many points spatially close compared to the image overall density.}
Our work focuses on integrating this naive model into the training loop of an object detector to guide the feature extraction, which was not considered in previous studies. Unlike~\cite{ciocarlan2023deep}, whose naive model is suitable for pixel-level classification (i.e. segmentation), we consider a different model that directly applies at object level and is thus more relevant for NN with bounding box proposals. 

\subsection{Object detection methods}

Object detection is the task of detecting objects of interest within an image \addSylv{and} identifying their locations with bounding boxes. \addSylv{S}everal types of deep learning approaches \addSylv{have been proposed for such a task}~\cite{girshick2015fast, redmon2016you}. YOLO framework is the most widely used one as it leads to great performance in various applications, with low execution time. It is a single-stage algorithm that uses a single convolutional neural network to predict together bounding box coordinates, objectness and classification scores.
Concretely, it divides the image into a grid and predicts the probability \addSylv{(denoted as the objectness score)} \addSylv{for any given} grid cell \addSylv{to} contain an object and the bounding box coordinates of the object if it exists. One issue with \addSylv{the} early versions of YOLO is that they struggle in detecting small objects. Indeed, if the object to detect is too small, it may only occupy a small portion of a grid cell, making it difficult for YOLO to detect it \addSylv{accurately}. To address this issue, YOLOv3~\cite{redmon2018yolov3} introduced a feature pyramid network (FPN) that combines \addSylv{the} features \addSylv{detected} at multiple scales. 
Some of the latest versions of YOLO, such as YOLOR~\cite{wang2021you} or YOLOv7~\cite{wang2022yolov7}, lead to competitive detection performance on several famous computer vision benchmarks, while also improving the execution speed. Tiny versions of YOLO with less convolutional layers \addSylv{have also been} proposed.

\section{Method}

\begin{figure*}[ht]
    \centering
    \includegraphics[width=17.8cm]{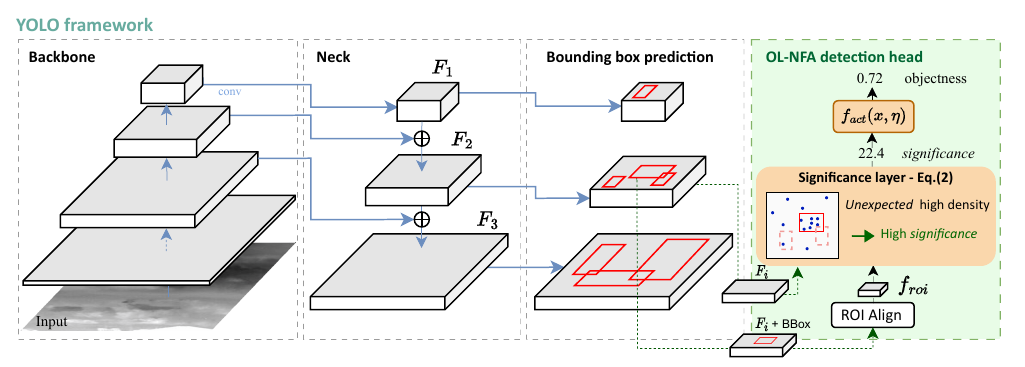}
    \caption{Integration of our OL-NFA detection head into a YOLO framework. This module can be added on top of any YOLO.}
    \label{fig:yolo_nfaobj}
\end{figure*}



\subsection{Overall architecture}

We propose a novel YOLO detection head, called OL-NFA for object-level NFA detection head, that integrates an \textit{a contrario} criterion to detect objects with features that \textit{unexpectedly} deviate from the background distribution. 
\addSylv{Our OL-NFA will compute objectness score based on NFA criterion, Eq.~\eqref{eq:nfa}, applied to feature maps derived by the network.}

The overall architecture of our approach is illustrated in Fig.~\ref{fig:yolo_nfaobj}. The infrared input images first go through a YOLO backbone that extracts feature maps at different scales. Then, the three lower-level features are combined together through the neck, which gives us the final feature maps \addSylv{$F_i$}
used to perform the detection at three levels\addSylv{:} $i \in \{1,2,3\}$. To achieve the detection, the bounding box coordinates are first predicted through a dense layer. We then introduce our OL-NFA module to re-estimate the objectness score for each bounding box using \addSylv{NFA} criterion. To do so, we extract $\eta$ regions of interest (ROI), denoted as $f_{roi}$, using ROI Align from Faster R-CNN~\cite{girshick2015fast}, and we compute a \textit{significance} score for each ROI through the significance layer described in Section~\ref{significance_layer}. Finally, these scores are ranged in $[0,1]$ via the function $f_{act}$ defined in in Section~\ref{significance_layer}, which allows us to apply the Binary Cross Entropy loss used in YOLO. 

\subsection{Differentiable integration of the \textit{a contrario} criterion}

\label{significance_layer}

Our significance layer in Fig.~\ref{fig:yolo_nfaobj} integrates the \textit{a contrario} criterion given in Eq.~\eqref{eq:nfa}. However, since this equation is (i) designed for binary images rather than greyscale feature maps, and (ii) not differentiable, several approximations were made in order to allow its integration into the YOLO training loop. The first difficulty raised by Eq.~\eqref{eq:nfa} is to count the number of ``true'' pixels $\kappa$ in $f_{roi} \in \mathbb{R}^2$. \addSylv{Thresholding} $f_{roi} $ to binarize it would break the back-propagation loop. \addSylv{Thus, w}e propose instead to \addSylv{consider real number membership coefficients (in the spirit of fuzzy clustering or classification), which boils down to handling}, for each pixel, a coefficient indicating the degree to which it belongs to \addSylv{the} set containing pixels with a value of $1$ in the binary case. For this purpose, we apply the sigmoid function $\sigma$ on the pixel \addSylv{values}, which allows us to \addSylv{approximate} the number of pixels contained in $f_{roi}$ for estimating the local density\addSylv{, by the sum of these fuzzy belonging coefficients}. The same approximation is made to compute the total number of points in $F_i$ for estimating the parameter $p$ (representing the global density of $F_i$) of the binomial law in Eq.~\eqref{eq:nfa}. The second issue is that the $\NFA$ function is discontinuous, non differentiable and, as we deal with objects having a small area $\nu$, it only takes very few distinct values. These elements make it difficult to integrate Eq.~\eqref{eq:nfa} ``as is'' into the training loop, with a working back-propagation. We therefore define the \textit{significance} $S\left(\kappa,\nu,p\right) = - \ln(\NFA(\kappa,\nu,p))$ and use the Hoeffding approximation when $\frac{\kappa}{\nu}>p$, leading to 
\begin{equation}
\label{eq:NFA_signif}
S\left(\kappa,\nu,p\right)\approx
\nu\left[\frac{\kappa}{\nu}\ln\left(\frac{\frac{\kappa}{\nu}}{p}\right)+\left(1-\frac{\kappa}{\nu}\right)\ln\left(\frac{1-\frac{\kappa}{\nu}}{1-p}\right)\right]-\ln \eta.
\end{equation}
This allows us to expand the domain of the function $S\left(\kappa,\nu,p\right)$ to $\mathbb{R}^3$, and to output more intermediate values. In the case of $\frac{\kappa}{\nu}\leq p$, we simply assign $\left(\kappa,\nu,p\right)=-\ln \eta$ as it corresponds to obvious background values. \addSylv{Finally, since} the \textit{significance} values range from $[-\ln(N_{test}),+\infty)$, where large values correspond to possible targets\addSylv{, t}o obtain objectness scores that range in $[0,1]$, we apply an asymmetric activation function $f_{act}(x, \eta)=2 \sigma (x+\ln \eta) -1$, with $x \in \mathbb{R}$ and $\eta \in \mathbb{N^*}$.


\section{Experiments}

\subsection{Dataset and metrics}
\label{datasets}

We evaluate our method on the NUAA-SIRST dataset~\cite{dai2021asymmetric}, which is one of the few infrared small target datasets publicly available and widely used in the literature. It is composed of $427$ infrared images, with wavelengths ranging from $950$ to $1200$ nm. Targets from NUAA-SIRST have a spatial extent that vary from $2-3$ pixels to more than $100$ pixels for the largest targets, which makes this dataset suitable to evaluate our method on a wide range of target sizes. Targets are drowned into challenging scenes such as textured clouds, as shown on the first row of Fig.~\ref{fig:sirst_viz}. We split the dataset into training, validation and test sets \addmoi{using} a ratio of $60:20:20$. We also evaluate the benefits of our method in a few-shot setting, by training the NN on $15$ and $25$ images only.
Regarding \addSylv{quantitative} evaluation, we focus on conventional detection metrics: the F1-score (F1) and the Average Precision (AP, area under Precision-Recall curve). We also rely on the Precision (Prec.) and the Recall (Rec.) to \addSylv{understand the achieved values of} F1-score. 
In the tables, the presented results have been averaged over three distinct training sessions, and the  standard deviation is given for F1 et AP in superscript. 
\subsection{Settings}

We add our OL-NFA detection head on top of YOLOv7-tiny, as this baseline has shown to lead to good performance on NUAA-SIRST dataset compared to other YOLO backbones. We compare it to several baselines\footnote{For YOLO baselines, we used the official PyTorch implementation of YOLO 
\href{https://github.com/WongKinYiu/yolov7}{WongKinYiu/yolov7}. For IRSTD baselines we used the implementation given by \href{https://github.com/kourenke/Review-Infrared-small-target-segmentation-networks}{kourenke/Review-Infrared-small-target-segmentation-networks}}: 1) segmentation networks specifically designed for IRSTD, namely ACM~\cite{dai2021asymmetric}, LSPM~\cite{huang2021infrared} and DNANet~\cite{li2022dense}; 2) YOLO baselines such as YOLOv3~\cite{redmon2018yolov3}, YOLOR~\cite{wang2021you}, YOLOv7 and YOLOv7-tiny~\cite{wang2022yolov7}. For the IRSTD segmentation NN, we use the training settings recommended in the original papers. All object detection NN are trained from scratch on Nvidia RTX6000 GPU for $600$ epochs, with Adam optimizer~\cite{kingma2014adam}, a batch size \addSylv{equal to} $16$ and a learning rate \addSylv{equal to} $0.001$. The same settings are used for the few-shot training. 

\subsection{\textit{A contrario} reasoning improves YOLO-based IRSTD}

\bgroup
\def\arraystretch{1.15}
\begin{table}[ht] 
\small
 \caption{Object-level \addSylv{metrics (}F1, AP, Prec., Rec.\addSylv{)} achieved by the compared methods on NUAA-SIRST. Best results are in bold and second best results are underlined. The inference time (frames per second, FPS) is also given. }
  \begin{tabular}{|c|cc|cc|c|} 
  \hline
    \textbf{Method} & \textbf{F1 }  & \textbf{AP} & \textbf{Prec. } & \textbf{Rec. }&\textbf{FPS}  \\
    \hline
     \multicolumn{6}{|l|}{Segmentation networks for IRSTD}\\ 
   \hline
  
   
   ACM~\cite{dai2021asymmetric} & $ 95.4$\textsuperscript{$\pm 1.7$}  & $ 95.2 ^{\pm 3.8} $  & $95.1 $ & $ 95.8 $ & 251 \\
   LSPM~\cite{huang2021infrared} & $ 85.0 ^{\pm 2.9} $  & $ 90.2 ^{\pm 0.8}$  & $86.6 $ & $ 83.5 $ & 125 \\
   DNANet~\cite{li2022dense} & $ \underline{96.9} ^{\pm 0.5} $  & $ \underline{98.1}^{ \pm 1.2}$  & $96.6$ & $ \textbf{97.2} $ & 33 \\
   \hline
     \multicolumn{6}{|l|}{Object detection methods}\\
   \hline
    
    YOLOv3~\cite{redmon2018yolov3} & $96.1^{\pm 0.3}$ & $97.5^{ \pm 0.1}$   &  $96.9 $  &   $95.4 $ & 144 \\
    
   YOLOR~\cite{wang2021you} & $95.7^{\pm 2.2}$ & $96.7 ^{\pm 1.1}$   &  $96.5 $  &   $94.9 $ & 136 \\
    
    YOLOv7~\cite{wang2022yolov7} & $96.5^{\pm 1.2}$ & $97.6 ^{\pm 0.7}$   &  $97.2 $  &   $95.9 $ & 147 \\
    
   YOLOv7-tiny & $96.5^{\pm 0.6}$ & $97.8 ^{\pm 0.4}$   &  $96.9 $  &   $\underline{96.2} $ & 256 \\
   \hline
   \textbf{Ours} & $\textbf{97.2} ^{\pm 0.6}$ & $\textbf{98.2} ^{\pm 0.2}$  &  $ \textbf{98.6} $ & $95.9 $  & 208 \\
   \hline
  \end{tabular}
  \label{res_SIRST}
\end{table}
\egroup

\begin{figure}[ht]
    \centering
    \includegraphics[width=8.9cm]{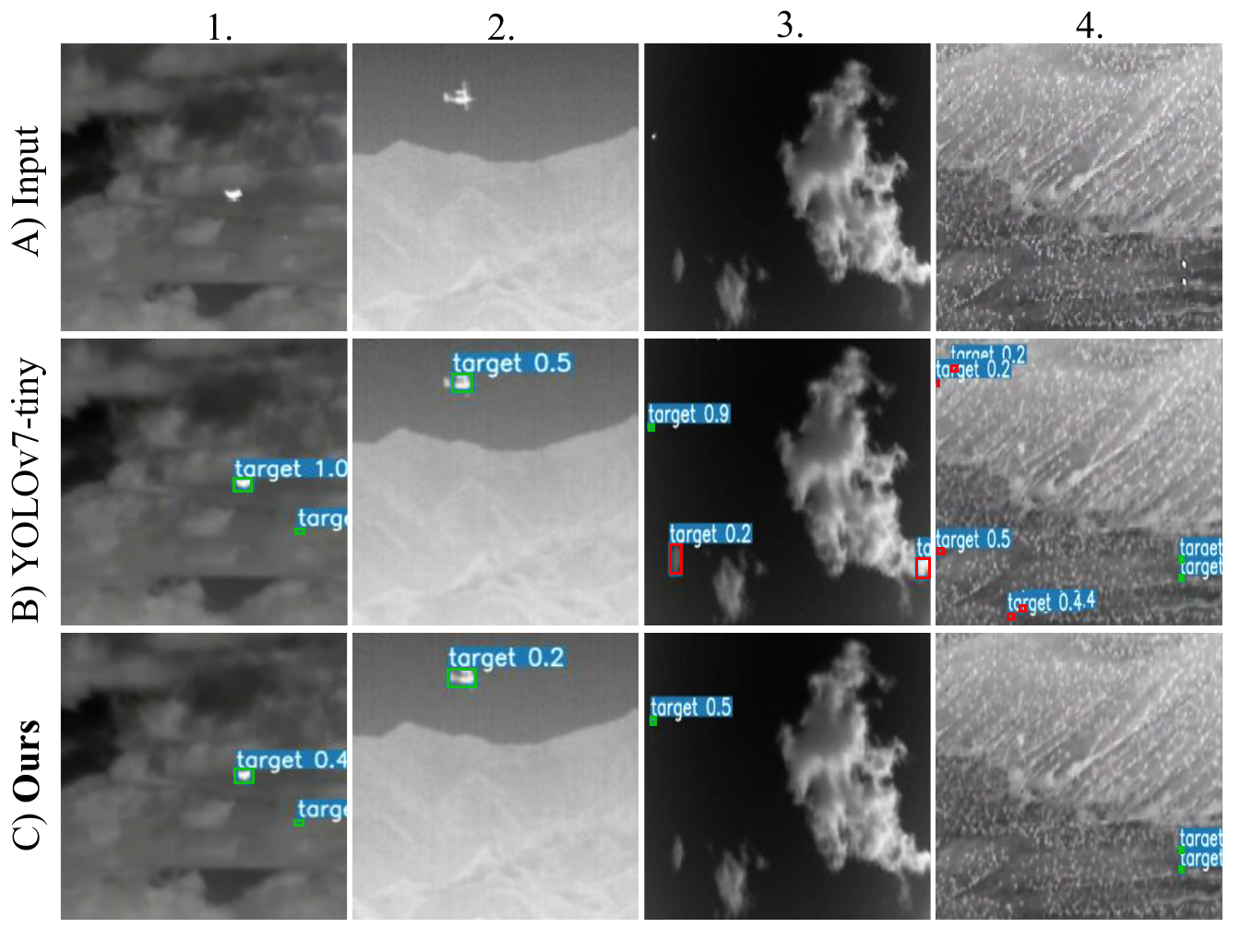}
    \caption{Qualitative results obtained with YOLOv7-tiny and our method on NUAA-SIRST dataset. Good detections and false positives are framed in green and red\addSylv{,} respectively.}
    \label{fig:sirst_viz}
\end{figure}

Table~\ref{res_SIRST} shows the performance \addSylv{achieved by each of} the compared methods on NUAA-SIRST. We can see that substituting conventional YOLO detection head with our OL-NFA not only improves YOLO for tiny object detection, but also bridges the performance gap observed between SOTA IRSTD segmentation NN and conventional object detection NN. \addSylv{Specifically}, our method achieves a F1 score higher by $0.7\%$ than the best YOLO baseline. The AP criterion is also increased by $0.4\%$. Moreover, our method performs slightly better in terms of F1 and AP than DNANet, which is SOTA method for IRSTD. The inference time for our method is also much lower than for DNANet, thus allowing for real-time object detection. The high performance of our OL-NFA module is mostly due to a higher precision with a limited loss in recall, which is explained by the NFA property of controlling the number of false alarms. Indeed, adding an \textit{a contrario} decision criterion helps in enhancing small object features and thus discriminating them from complex backgrounds. This can be seen in Fig.~\ref{fig:sirst_viz}, where the best YOLO baseline leads to several false alarms for inputs 3 and 4, while our method provides correct detections without any false alarm. 

\subsection{OL-NFA brings robustness towards few-shot settings}
\bgroup
\def\arraystretch{1.15}
\begin{table}[h] 
\small
 \caption{Results achieved in 15 and 25-shot settings on NUAA-SIRST. Best results are in bold. }
  \begin{tabular}{|c|c|c|c|c|} 
  \hline
    \multirow{2}{*}{\textbf{Method}} & \multicolumn{2}{|c|}{\textbf{15-shots}}& \multicolumn{2}{|c|}{\textbf{25-shots}}\\\cline{2-5}
    
   & \textbf{F1 }  & \textbf{AP }  &\textbf{F1 }  & \textbf{AP }\\
   \hline
  YOLOv7-tiny& $50.7^{\pm 7.0}$ & $51.3 ^{\pm 7.0}$& $68.0^{\pm 6.6}$ & $69.6 ^{\pm 8.4}$   \\
   \hline
\textbf{Ours}    & $ \textbf{85.0} ^{\pm 5.0}$ & $ \textbf{90.5} ^{\pm 5.2}$  & $ \textbf{89.7} ^{\pm 4.2}$ & $ \textbf{93.4} ^{\pm 2.0}$ \\
   \hline
  \end{tabular}
  \label{res_SIRST_25shot}
\end{table}
\egroup
    
  

 One important motivation of integrating \textit{a contrario} reasoning into a NN is that the network learns to discriminate small targets by learning a representation of background elements rather than the targets themselves. It should thus provide robustness to the NN towards weak training conditions. To confirm our intuition, we quantitatively evaluate the benefit of the proposed approach in a few-shot setting on NUAA-SIRST dataset. For this purpose, we trained the networks on $15$ and $25$ images. For each few-shot setting, we train the detectors on three distinct folds, with no overlap between them. The results obtained on the test set defined in Section~\ref{datasets} are averaged over these three folds and \addSylv{computed means} are given in Table~\ref{res_SIRST_25shot}. It can be seen that our method performs significantly better in a frugal setting than the baseline. Indeed, in those cases, both F1 score and Average Precision are increased by at least $20\%$. \addSylv{We thus conclude that a}dding an object-level NFA to the baseline significantly improves its robustness towards frugal setting: the F1 score is decreased by only $15\%$ when dividing by more than $10$ the number of training samples and the AP is maintained above $90\%$.

\section{Conclusion}
In this paper, we propose a novel YOLO detection head named OL-NFA that integrates an \textit{a contrario} decision criterion into the training loop of a YOLO network. It forces the network to model the background distribution rather than the objects to detect. Extensive experiments have shown that our method not only significantly improves YOLO performance for small object detection in frugal and few-shot settings, but also performs on par with SOTA segmentation networks for small target detection. This promising performance encourages to consider further research into using \textit{a contrario} paradigm for tiny object detection.

\bibliographystyle{IEEEbib}
\bibliography{strings,refs}

\end{document}